\documentclass[10pt,twocolumn,letterpaper]{article}

\usepackage[pagenumbers]{cvpr} 

\usepackage[dvipsnames]{xcolor}

\definecolor{cvprblue}{rgb}{0.21,0.49,0.74}
\usepackage[pagebackref,breaklinks,colorlinks,citecolor=cvprblue]{hyperref}
\usepackage{multirow}
\usepackage{natbib}

\def\paperID{*****} 
\def\confName{CVPR}
\def\confYear{2024}

\title{Semantic Segmentation on VSPW Dataset through Masked Video Consistency}

\author{
Chen Liang$^\text{1,2}$\thanks{Equal contribution.} ~~~ Qiang Guo$^\text{1}$\footnotemark[1] ~~~ Chongkai Yu$^\text{1}$ ~~~ Chengjing Wu$^\text{1}$ ~~~ Ting Liu$^\text{1}$ ~~~ Luoqi Liu$^\text{1}$ \\
$^\text{1}$MT Lab, Meitu Inc.\\
$^\text{2}$Chinese Academy of Sciences, Institute of Automation\\
{\tt\small \{lc25, gq5, yck, ethan, lt, llq5\}@meitu.com}
}

\begin{document}

\maketitle

\begin{abstract}
Pixel-level Video Understanding requires effectively integrating three-dimensional data in both spatial and temporal dimensions to learn accurate and stable semantic information from continuous frames. However, existing advanced models on the VSPW dataset have not fully modeled spatiotemporal relationships. In this paper, we present our solution for the PVUW competition, where we introduce masked video consistency (MVC) based on existing models. MVC enforces the consistency between predictions of masked frames where random patches are withheld. The model needs to learn the segmentation results of the masked parts through the context of images and the relationship between preceding and succeeding frames of the video. Additionally, we employed test-time augmentation, model aggeregation and a multimodal model-based post-processing method. Our approach achieves 67.27\% mIoU performance on the VSPW dataset, ranking 2nd place in the PVUW2024 challenge VSS track.
\end{abstract}    
\section{Introduction}
\label{sec:intro}

The PVUW2024 VSS Track focuses on the Pixel-level Video Understanding in the Wild \citep{miao2021vspw, Miao_2022_CVPR}, which is a critical task in advancing computer vision. Unlike traditional image segmentation, this challenge addresses the need for video segmentation, recognizing that real-world applications are video-based rather than static. The goal is to assign predefined semantic labels to every pixel in all frames of a given video, leveraging temporal information to improve predictive accuracy. This task is more challenging than image-based segmentation since it requires the integration of spatial and temporal data to produce accurate and stable semantic information across continuous frames. The final evaluation is conducted on the VSPW \citep{miao2021vspw} dataset, with using mean Intersection over Union (mIoU) and Video Consistency (VC) as evalutation metrics.

However, existing methods \citep{zhang2023dvis, zhang2023dvis++} face several challenges: they struggle with finely segmenting object edges, often misclassifying regions of the same object with nearly unchanged visual features in the original image as different categories. Additionally, their capability for temporal modeling remains underdeveloped, resulting in the model's inability to maintain stable segmentation of the same semantic category across adjacent frames. To address these issues, we introduce Masked Video Consistency (MVC). MVC optimizes existing models by enforcing consistency constraints in both temporal and spatial dimensions. It encourages the model to maintain consistent predictions for masked frames, where random patches are hidden, by leveraging the contextual information from surrounding frames and the spatial structure of the video. This dual-focus on spatial and temporal consistency aims to overcome the limitations of current models and achieve more accurate and stable video semantic segmentation. 

To alleviate the burden during the model training process, we employed a Test-Time Augmentation (TTA) strategy that integrates segmentation results from various input modifications using a single model, rather than assembling multiple models. Specifically, we applied different input scales and image flips. This approach allows us to leverage the strengths of multiple augmented views of the input data, enhancing the robustness and accuracy of the segmentation results without the additional computational overhead of training and maintaining multiple models. By using TTA, we ensure that our model can generalize better to diverse input conditions, leading to more reliable performance in practical applications.

In the post-processing stage, we utilized the large multimodal model \citep{liu2023improved, liu2024visual, achiam2023gpt} to correct some of the video segmentation results. During our observation of the final segmentation outcomes, we found that the model struggled to maintain consistency in segmenting challenging stuff classes across the entire video sequence. Additionally, it sometimes misclassified a single class into multiple similar class regions within a single frame. We attributed this issue to the model's lack of sufficient training data. To address this, we employed a large multimodal model to refine these results, ultimately ensuring consistent segmentation throughout the entire video.

By combining the above strategies, we achieved a 67.27\% mIoU on the final test set. Additionally, VC$_8$ and VC$_{16}$ scores were 94.99\% and 93.12\%, respectively, demonstrating the consistency and stability of our segmentation results.

\section{The proposed method}

\subsection{Baseline Model}

DVIS++ \citep{zhang2023dvis++} is a general-purpose video segmentation model that has achieved state-of-the-art performance across various benchmarks, including video instance segmentation, video semantic segmentation, and video panoptic segmentation. Therefore, we chose DVIS++ as the baseline model for our experiments and explored its capabilities further. Unlike previous models with end-to-end training, DVIS++ decouples video segmentation into three cascaded sub-tasks: segmentation, tracking, and refinement. This task-decoupling design allows us to separately explore the temporal and spatial consistency brought by Masked Video Consistency (MVC).

\subsection{Masked Video Consistency}

To accurately identify an object or region, a model typically needs to utilize clues from different parts of an image. These clues can come from the local information of the same image block corresponding to the units in the feature map, or from the surrounding image blocks, corresponding to the contextual information of the image \citep{hoyer2019grid}. Additionally, for temporal modeling tasks such as video segmentation, the image information from other video frames also constitutes a crucial temporal context.

Many network architectures \citep{he2016deep, dosovitskiy2020image} have the capability to integrate local and contextual information into their features. However, they struggle to effectively enforce the learning of useful context clues. Our experiments have shown that while DVIS++ models inter-frame relationships within a unified video through the Referring Tracker and Temporal Refiner, it still exhibits instability in the final segmentation results. To inject more contextual clues into the model, we propose the plug-and-play Masked Video Consistency (MVC) approach.

Specifically, for each image, a patck Mask $\mathcal{M}$ is obtained by

\begin{equation}
\mathcal{M}_{ij} = \left[ v > r \right] \quad \text{with} \quad v \sim \mathcal{U}(0, 1)
\end{equation}

where $[\cdot]$ denotes the Iverson bracket, $b$ is the patch size, $r$ is the mask ratio, $i$ and $j$ represent the positions of image blocks. The masked image is obtained by element-wise multiplication of mask and image

\begin{equation}
    x^M = \mathcal{M} \odot x^T
\end{equation}

The final prediction $\hat{y}^{M}$ is inferred from the images that were removed from a portion of the mask

\begin{equation}
    \hat{y}^{M} = f_{\theta}(x^M)
\end{equation}

The final loss function obtained for model training is as follows

\begin{equation}
    \underset{\theta}{min}\frac{1}{N}\sum_{k=1}^{N}(\mathcal{L}_k + \lambda\mathcal{L}_k^M)
\end{equation}

The loss $\mathcal{L}$ is defined according to the baseline model. 

\subsection{Model aggregation}

Furthermore, we combined the segmentation results of two models: one trained with the Masked Video Consistency (MVC) strategy and one trained without it. This dual-model approach aims to mitigate the potential instability introduced by MVC when there are significant differences between consecutive video frames (e.g., non-continuous frames). We leverage optical flow to measure the temporal consistency of the segmentation maps.

Given two consecutive frames from a video, $x_t$ and $x_{t+1}$, we compute the optical flow using Farneback method \citep{farneback2003two}. 

\begin{equation}
    \mathbf{F}_{t \rightarrow t+1} = \text{OF}(x_{t}, x_{t+1})
\end{equation}

where $\mathbf{F}_{t \rightarrow t+1}$ represents the optical flow from $x_t$ to $x_{t+1}$, $\text{OF}$ is the Farneback method \citep{farneback2003two}.

With obtaining optical flow $\mathbf{F}_{t \rightarrow t+1}$, we then warp the segmentation result $y_t$ to generate the prediction of frame $x_{t+1}$

\begin{equation}
    \tilde{y}_{t+1} = \text{Warp}(y_{t}, \mathbf{F}_{t \rightarrow t+1})
\end{equation}

To evaluate the quality of the segmentation, we compare the warped segmentation map $\tilde{y}_{t+1}$ with $y_{t+1}$ using the Structural Similarity Index (SSIM)

\begin{align}
\begin{split}
    \text{SSIM}(\tilde{y}_{t+1}, y_{t+1}) = \hspace{15em}\\
    \frac{(2\mu_{\tilde{y}_{t+1}}\mu_{y_{t+1}} + C_1)(2\sigma_{\tilde{y}_{t+1}y_{t+1}} + C_2)}{(\mu_{\tilde{y}_{t+1}}^2 + \mu_{y_{t+1}}^2 + C_1)(\sigma_{\tilde{y}_{t+1}}^2 + \sigma_{y_{t+1}}^2 + C_2)}
\end{split}
\end{align}

where \(\mu\) and \(\sigma\) represent the mean and variance of the respective segmentation maps, and \(C_1\) and \(C_2\) are constants to stabilize the division.

We repeat the above process for both sets of segmentation results. The segmentation result with the higher SSIM score is considered to be temporally more consistent and, thus, of higher quality. We select each video based on this strategy for the segmentation results of the two models, and ultimately obtain the integration results of the two models.

\subsection{Test-time augmentation}

We utilized test-time augmentation to integrate segmentation results from the same model with different inputs, including stochastic flipping and multi-scale data enhancement. The scales used in multi-scale inference include {0.9, 1.0, 1.1, 1.2}, and each scale involves horizontal flipping.

After obtaining multiple model predictions, we integrated the results using a voting approach. For each pixel in each image, we perform voting across all prediction results. In the event of a tie, we prioritize the result corresponding to the higher resolution input, as high-resolution inputs potentially contain more image details, which facilitates model analysis and inference. 

For each pixel $j$, the final prediction result $V_j$ can be expressed as:

\begin{equation}
V_j = \begin{cases} 
\text{mode}(\{y_{i,j} \mid i = 1, 2, \ldots, N\}) & \text{if no tie} \\
y_{i^*, j} & \text{if tie}
\end{cases}
\end{equation}

where the $\text{mode}$ function calculates the most frequent class in the set, $N$ represents the total number of different augmentations, and $i^*$ is the index corresponding to the high-resolution input in the augmentation set. By TTA, the reliability of the final results is further enhanced.

\subsection{Post-processing}

Although the aforementioned method enhances the model's ability to utilize existing information, such as learning the patterns corresponding to each category from the training set images and learning the inter-frame correspondence in video sequences, it is still limited by the finite scenes in the VSPW training set. As a result, the model struggles to generalize to unseen scenes. In our experiments, we observed that while the model produced relatively complete segmentation masks for some challenging scenes, it tended to make classification errors in the masks for categories that could be ambiguous. To address these issues, we attempted to use a large multimodal model \citep{liu2023improved, liu2024visual, achiam2023gpt} to post-process the segmentation results, leveraging its extensive and rich training samples.
\section{Experiments}

\subsection{Datasets and evaluation metrics}

VSPW \citep{miao2021vspw} is a dataset with the target of advancing the scene parsing task for images to videos. VSPW contains 3,536 videos, including 251,633 frames from 124 categories. Each video contains a well-trimmed long-temporal shot, lasting around 5 seconds on average. 

We use Mean IoU (mIoU) to evaluate the segmentation performance. To evaluate the stability of predictions, we also provide experimental results of Video Consistency (VC). The evaluation metrics are consistent with the leaderboard metric of the competition. 

\subsection{Implementation details}

We use DVIS++ \citep{zhang2023dvis++} as our baseline model, which emplys Mask2Former \citep{cheng2022masked} as the segmenter. The referring tracker utilizes six transformer denoising blocks, and the temporal refiner emplys six temporal decoder blocks. We use the open-source Segmenter weights of DVIS++ for finetune and retrain the following modules with the same configuration. All experiments were conducted on eight NVIDIA V100 GPUs.

\subsection{Ablation studies}

\paragraph{Ablation study of MVC. }

To validate the effectiveness of the proposed Masked Video Consistency (MVC) in improving model segmentation capabilities, we conducted tests on DVIS++ using different backbones at various training stages in Tab. \ref{tab:mvc}. For comparison with previous models, we trained using only the training set and calculated all evaluation metrics on the validation set. The experimental results clearly demonstrate that MVC enhances segmentation performance (mIoU) and video segmentation consistency (VC), outperforming the state-of-the-art DVIS++ (Offline) with 0.5\% mIoU, 0.3\% mVC, and 0.3\% mVC. Additionally, we validated the effectiveness of MVC across different stages of DVIS++, achieving improvements of 0.6\% mIoU, 0.7\% mVC, and 0.9\% mVC for the Online model as well. Notably, our approach does not introduce additional model parameters or increase training memory overhead; instead, it provides additional training cues based on the existing model.

\begin{table}[ht]
\centering
\resizebox{\columnwidth}{!}{
\begin{tabular}{l|l|ccc}
\toprule \toprule
\textbf{Method} & \textbf{Backbone} & \textbf{mVC$_8$} & \textbf{mVC$_{16}$} & \textbf{mIOU} \\ \midrule
Mask2Former \citep{cheng2022masked} & ResNet-50 & 87.5 & 82.5 & 38.4 \\
Video-kMax \citep{shin2024video} & ResNet-50 & 86.0 & 81.4 & 44.3 \\
Tube-Link \citep{li2023tube} & ResNet-50 & 89.2 & 85.4 & 43.4 \\
MPVSS \citep{weng2024mask} & ResNet-50 & 84.1 & 77.2 & 37.5 \\
DVIS(online) \citep{zhang2023dvis} & ResNet-50 & 92.0 & 90.9 & 46.6 \\
DVIS(offline) \citep{zhang2023dvis} & ResNet-50 & 93.2 & 92.3 & 47.2 \\
DVIS++(online) \citep{zhang2023dvis++} & ResNet-50 & 92.3 & 91.1 & 46.9 \\
DVIS++(offline) \citep{zhang2023dvis++} & ResNet-50 & 93.4 & 92.4 & 48.6 \\ 
\midrule
DeepLabv3+ \citep{chen2018encoder} & ResNet-101 & 83.5 & 78.4 & 35.7 \\
TCB \citep{miao2021vspw} & ResNet-101 & 86.9 & 82.1 & 37.5 \\
Video K-Net \citep{li2022video} & ResNet-101 & 87.2 & 82.3 & 38.0 \\
MRCFA \citep{sun2022mining} & MiT-B2 & 90.9 & 87.4 & 49.9 \\
CFFM \citep{sun2022coarse} & MiT-B5 & 90.8 & 87.1 & 49.3 \\
Video K-Net+ \citep{li2022video} & ConvNeXt-L & 90.1 & 87.8 & 57.2 \\
Video kMax \citep{shin2024video} & ConvNeXt-L & 91.8 & 88.6 & 63.6 \\
TubeFormer \citep{kim2022tubeformer} & Axial-ResNet-50 & 92.1 & 88.0 & 63.2 \\
MPVSS \citep{weng2024mask} & Swin-L & 89.6 & 85.8 & 53.9 \\
DVIS(online) \citep{zhang2023dvis} & Swin-L & 95.0 & 94.3 & 61.3 \\
DVIS(offline) \citep{zhang2023dvis} & Swin-L & 95.1 & 94.4 & 63.3 \\
DVIS++(online) \citep{zhang2023dvis++} & ViT-L & 95.0 & 94.2 & 62.8 \\
DVIS++(offline) \citep{zhang2023dvis++} & ViT-L & 95.7 & 95.1 & 63.8 \\ 
DVIS-M++(online) & ViT-L & 95.7 & 95.1 & 63.6 \\
DVIS-M++(offline) & ViT-L & \textbf{96.0} & \textbf{95.4} & \textbf{64.3} \\ 
\bottomrule \bottomrule
\end{tabular}}
\caption{\textbf{Comparison of different methods on the VSPW validation set.} mVC$_k$ means that a clip with $k$ frames is used. The best results are highlighted in bold.}
\label{tab:mvc}
\end{table}

\paragraph{Ablation study of extra training data. }

To fully utilize the existing annotated data, we expanded the original training set using the VSPW \citep{miao2021vspw} open-source validation set data. Specifically, due to the difficulty in reproducing the pre-trained weights, we fine-tuned the Segmenter weights provided by the official DVIS++ repository \citep{zhang2023dvis++} on the combined training and validation sets. Subsequently, we re-trained the Reference Tracker and Temporal Refiner using the fine-tuned Segmenter weights. The experimental results after augmenting the training data are shown in Tab. \ref{tab:extra}. As seen in the table, whether or not the MVC training strategy was employed, the inclusion of the validation set provided the model with additional semantic knowledge, ultimately leading to better segmentation performance on the test set.

\begin{table}[ht]
\centering
\resizebox{\columnwidth}{!}{
\begin{tabular}{l|l|cc|c}
\toprule \toprule
\multicolumn{1}{l|}{\multirow{2}{*}{\textbf{Method}}} & \multicolumn{1}{l|}{\multirow{2}{*}{\textbf{Backbone}}} & \multicolumn{2}{c|}{\textbf{Dataset used}}          & \textbf{mIoU} \\
\multicolumn{1}{l|}{}                        & \multicolumn{1}{l|}{}                          & training & \multicolumn{1}{l|}{validation} &      \\ \midrule
DVIS++(offline) & ViT-L & $\checkmark$ &  & 61.6 \\
DVIS++(offline) & ViT-L & $\checkmark$ & $\checkmark$ & 62.7 \\
DVIS-M++(offline) & ViT-L & $\checkmark$ &  & 62.2 \\
DVIS-M++(offline) & ViT-L & $\checkmark$ & $\checkmark$ & \textbf{63.5} \\
\bottomrule \bottomrule
\end{tabular}}
\caption{\textbf{Ablation study of extra training data.} All results are on the test set in the final state. }
\label{tab:extra}
\end{table}

\paragraph{Ablation study of test-time augmentation. }

The default test crop size for the DVIS++ model \citep{zhang2023dvis++} is 720p. To evaluate whether different crop sizes and image flipping provide more effective input information for the model, we conducted separate experiments on these two TTA methods. The experimental results are shown in Tab. \ref{tab:tta}. Without flipping, using four input resolutions yielded better performance than using a single resolution. Additionally, images with reduced crop sizes also contributed to the model's final voting results. On this basis, flipping the image can further improve the prediction results. We believe that higher resolutions help the model to segment image and video details more precisely, while lower resolutions assist the model in ignoring some class-irrelevant texture information. This also helps prevent the model from segmenting a single class area into multiple different class blocks to a certain extent.

\begin{table}[ht]
\centering
\begin{tabular}{l|c|c}
\toprule \toprule
\textbf{Scale}         & \textbf{Flip}  & \textbf{mIoU} \\ \midrule
\{1.0\}                &                & 63.5          \\
\{1.0\}                & $\checkmark$   & 63.6          \\
\{1.0, 1.1, 1.2\}      &                & 63.9          \\
\{1.0, 1.1, 1.2\}      & $\checkmark$   & 64.4          \\
\{0.9, 1.0, 1.1, 1.2\} &                & 64.1          \\
\{0.9, 1.0, 1.1, 1.2\} & $\checkmark$   & \textbf{64.4} \\ 
\bottomrule \bottomrule
\end{tabular}
\caption{\textbf{Ablation study of test-time augmentation.} All results are on the test set in the final state. }
\label{tab:tta}
\end{table}

\paragraph{Ablation study of model aggregation}

Based on the optical flow method, we evaluated the quality of segmentation results from different models on various videos. During the model integration phase, we selected the segmentation results with higher quality for integration. Tab. \ref{tab:inte} shows the improvement brought by model integration. Compared to DVIS-M++, the integrated model achieved a 1.4\% increase in mIoU.

\begin{table}[ht]
\centering
\begin{tabular}{c|c|c}
\toprule \toprule
\textbf{DVIS++} & \textbf{DVIS-M++} & \textbf{mIoU} \\ \midrule
$\checkmark$    &                   & 63.6          \\
                & $\checkmark$      & 64.4          \\
$\checkmark$    & $\checkmark$      & \textbf{65.8} \\ 
\bottomrule \bottomrule
\end{tabular}
\caption{\textbf{Ablation study of model aggregation.} All results are on the test set in the final state.}
\label{tab:inte}
\end{table}

\paragraph{Ablation study of post processing.} 

We use a large multimodal model to correct the existing segmentation results, leveraging the advantages of the extensive training data used for the multimodal model. Tab. \ref{tab:vlm} demonstrates the comparative performance of GPT-4 \citep{achiam2023gpt} and LLAVA \citep{liu2024visual} in predicting the class of objects in images. The prompt defined for this evaluation is \textit{"Is the \{stuff\} in the image a \{class1, class2, ...\}? Please give me the only answer."} While GPT-4 provides concise and accurate predictions, LLAVA misclassifies the water in both images as a river. GPT-4 showcases higher recognition accuracy in most of the cases. Given its superior performance in correctly classifying the objects, we adopted GPT-4 to refine segmentation results via a Q\&A approach. By leveraging GPT-4's accurate predictions, we ensure more reliable and precise segmentation corrections. 

\begin{table*}[t]
\centering
\begin{tabular}{|p{3cm}|p{7cm}|p{7cm}|}
\toprule \toprule
\multicolumn{3}{l}{\textbf{Category correction based on VLM}} \\ \midrule
\multicolumn{1}{l|}{} & \multicolumn{1}{p{7cm}|}{\includegraphics[width=7cm]{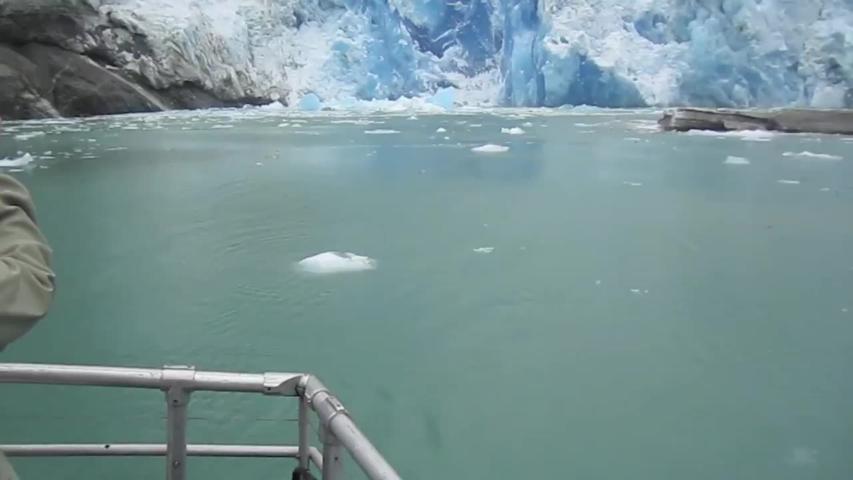}} & \multicolumn{1}{p{7cm}}{\includegraphics[width=7cm]{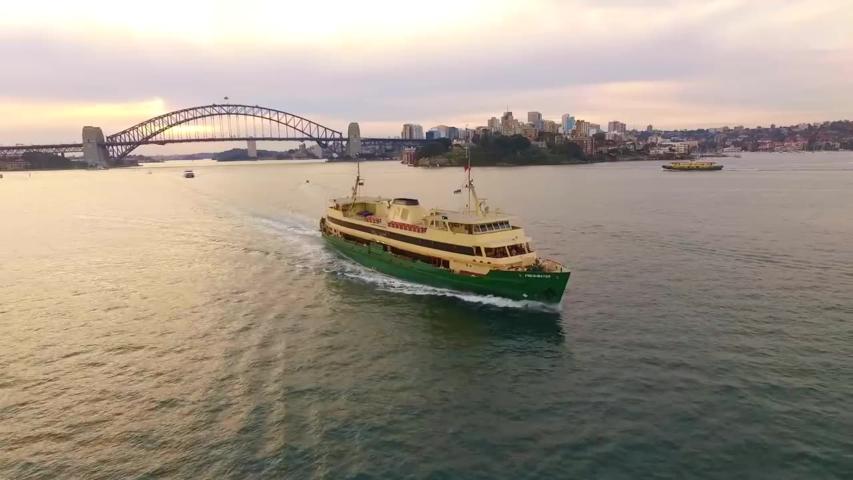}} \\ 
\multicolumn{1}{l|}{User}  & \multicolumn{1}{p{7cm}|}{Is the water in the image a \textbf{river}, \textbf{lake} or \textbf{sea}? Please give me the only answer.} & \multicolumn{1}{p{7cm}}{Is the water in the image a \textbf{river}, \textbf{lake} or \textbf{sea}? Please give me the only answer.} \\ \midrule
\multicolumn{1}{l|}{GPT-4} & \multicolumn{1}{p{7cm}|}{The water in the image is a \textcolor{blue}{\textbf{lake}}.} & \multicolumn{1}{p{7cm}}{The water in the image is a \textcolor{blue}{\textbf{sea}}.} \\ \midrule
\multicolumn{1}{l|}{LLAVA} & \multicolumn{1}{p{7cm}|}{The water in the image appears to be a \textbf{river}, as it is flowing and there is a visible waterfall, which is a common feature in rivers.} & \multicolumn{1}{p{7cm}}{The water in the image appears to be a \textbf{river}, as indicated by the presence of a bridge in the background, which is a common feature along rivers in urban areas. The bridge's design and the presence of buildings on the far bank also suggest that this is a river that runs through a city.} \\ \bottomrule \bottomrule
\end{tabular}
\caption{\textbf{Ablation study of different VLM models. }We compare the examples and results of using different VLM models to correct challenging segmentation results. GPT-4 has higher recognition accuracy compared to LLAVA.}
\label{tab:vlm}
\end{table*}

\subsection{Comparison with other teams}

In the final stage of the competition, our model achieved 67.3\% mIoU, securing the 2nd position among all participating teams. Meanwhile, the mVC$_8$ reached 95.0\%, earning the 1st place. Detailed results are presented in Tab. \ref{tab:board}.

\begin{table}[ht]
\centering
\begin{tabular}{l|c|c|c}
\toprule \toprule
\textbf{Team}   & \textbf{mIoU}     &\textbf{mVC$_8$}   & \textbf{mVC$_{16}$}   \\ \midrule
SiegeLion       & \textbf{67.8}     & \underline{94.8}  & 92.9                  \\
Ours            & \underline{67.3}  & \textbf{95.0}     & \underline{93.1}      \\
kevin1234       & 63.9              & \underline{94.8}  & \textbf{93.3}         \\ 
bai\_kai\_shui  & 63.8              & 94.6              & 92.9                  \\
JMCarrot        & 63.4              & 94.6              & 92.9                  \\
ipadvideo       & 58.5              & 90.7              & 88.0                  \\
\bottomrule \bottomrule
\end{tabular}
\caption{\textbf{Comparisons with other teams on the final state.} The \textbf{first} and \underline{second} highest scores are represented by bold font and underline respectively.}
\label{tab:board}
\end{table}

\section{Conclusion}

In this report, we present an innovative approach to enhance video segmentation performance. The Proposed training strategy Masked Video Consistency optimize the model's utilization of temporal and spatial contextual cues in video segmentation tasks. By integrating Test-Time Augmentation (TTA) strategies, model aggregation and utilizing large multimodal models for post-processing, we enhance the robustness and accuracy of segmentation results, and explore the potential of other domain models in segmentation tasks. As a result, we obtain a significant improvements in both mIoU and video consistency metrics and achieve the 2nd place in the PVUW 2024 VSS track. 
{
    \small
    \bibliographystyle{ieeenat_fullname}
    \bibliography{main}
}


\end{document}